\documentclass[10pt,twocolumn]{article}

\usepackage{cvpr}
\usepackage{times}

\usepackage{graphicx}
\usepackage{amsmath}
\usepackage{amssymb}
\usepackage{xcolor}
\usepackage{colortbl}
\usepackage{subcaption}
\usepackage{booktabs}
\usepackage{multirow}
\usepackage{array}
\usepackage{caption}
\usepackage{float}
\usepackage{enumitem}
\usepackage{placeins}  

\usepackage{pifont}
\newcommand{\xmark}{\ding{55}}%

\usepackage[pagebackref=true,breaklinks=true,colorlinks,bookmarks=false]{hyperref}

\cvprfinalcopy 

\def\cvprPaperID{****} 

\ifcvprfinal\fi
\begin{document}

\title{DragGANSpace: Latent Space Exploration and Control for GANs}

\author{Kirsten Odendaal\\ 
Georgia Institute of Technology\\
Atlanta, GA, USA\\
{\tt\small kodendaal3@gatech.edu}
\and
Neela Kaushik\\
Georgia Institute of Technology\\
Atlanta, GA, USA\\
{\tt\small nkaushik31@gatech.edu}
\and
Spencer Halverson\\
Georgia Institute of Technology\\
Atlanta, GA, USA\\
{\tt\small bhalverson8@gatech.edu}
}

\maketitle

\begin{abstract}

This work integrates StyleGAN, DragGAN and Principal Component Analysis (PCA) to enhance the latent space efficiency and controllability of GAN-generated images. StyleGAN provides a structured latent space, DragGAN enables intuitive image manipulation, and PCA reduces dimensionality and facilitates cross-model alignment for more streamlined and interpretable exploration of latent spaces. We apply our techniques to the Animal Faces High Quality (AFHQ) dataset, and find that our approach of integrating PCA-based dimensionality reduction with the DragGAN framework for image manipulation retains performance while improving optimization efficiency. Notably, introducing PCA into the latent W+ layers of DragGAN can consistently reduce the total optimization time while maintaining good visual quality and even boosting the Structural Similarity Index Measure (SSIM) of the optimized image, particularly in shallower latent spaces (W+ layers = 3). We also demonstrate capability for aligning images generated by two StyleGAN models trained on similar but distinct data domains (AFHQ-Dog and AFHQ-Cat), and show that we can control the latent space of these aligned images to manipulate the images in an intuitive and interpretable manner. Our findings highlight the possibility for efficient and interpretable latent space control for a wide range of image synthesis and editing applications.

\end{abstract}

\section{Introduction}

Generative Adversarial Networks (GANs) have significantly advanced the field of image synthesis, enabling the creation of highly realistic images across diverse domains. However, challenges remain in achieving precise control over their latent spaces, which are essential for both fine-grained image manipulation and the interpretability of the outputs. For example, if one desires to generate an image with a specific pose or expression, how can this be done? These challenges become even more pronounced when fine-tuning GANs on new datasets, as latent space attributes may not align with desired image modifications.

Our research makes use of StyleGAN2, a framework known for its high-quality and flexible latent space \cite{karras2020analyzingimprovingimagequality}.  To address the challenges of latent space manipulation, we focus on two main aspects. First, we investigate the benefits of applying Principle Component Analysis (PCA) to reduce dimensionality in latent space, to improve the computational efficiency of applying DragGAN \cite{Pan_2023}, an advanced but expensive image manipulation methodology.  Inspired by GANSpace \cite{härkönen2020ganspacediscoveringinterpretablegan}, PCA reduces the complexity of latent space, improving computational efficiency, and enabling streamlined latent space manipulation both within and between trained GANs. Second, we explore cross-model alignment, examining the extent to which the latent space retains semantic coherence between StyleGAN models trained on related but distinct datasets. This alignment aims to facilitate meaningful edits and ensure consistent latent space interpretability across multiple models. If successful, our approach will enable faster and more intuitive image editing, making advanced GAN methodologies like DragGAN accessible for broader applications in fields such as graphic design, virtual reality, and medical imaging. By ensuring semantic coherence between models, our findings could also enhance the transferability of GAN-based tools, streamlining workflows across diverse datasets and domains.

\subsection{Data}
We use the Animal Faces High-Quality (AFHQ) dataset, which comprises 15,000 high-quality images of cat, dog, and wildlife faces (equally distributed) at 512×512 resolution \cite{choi2020starganv2diverseimage}. For our experiments, we focus on the dog subset to investigate fine-tuning and PCA improvements in DragGAN, while using both the cat and dog subsets for the cross-model alignment analysis. This dataset presents a challenging baseline which is applicable to many real-world problems due to its multiple domain structure and the diversity of breeds within each domain, making it an interesting and fun case for our investigations.






\section{Technical Background}


\subsection{StyleGAN}
StyleGAN2, developed by NVIDIA, is a widely used GAN framework known for its structured and adaptable latent space, which enables high-quality image synthesis across various domains \cite{karras2020analyzingimprovingimagequality}. This flexible latent space structure allows for the adjustment of image style layers and attributes, giving users control over various aspects of generated images, such as pose, expression, and lighting. The core of StyleGAN’s success in controllability lies in its $W+$ latent space, a more expressive extension of the original latent space, where each layer of the generator can receive a unique latent code. This feature supports disentangled and precise manipulation of image attributes \cite{wu2020stylespaceanalysisdisentangledcontrols}.


Although StyleGAN and its successor's \cite{karras2019stylebasedgeneratorarchitecturegenerative,karras2020analyzingimprovingimagequality} latent space allows for improved control, manipulating these latent attributes can be complex and computationally intensive. This has led to the adoption of complementary methods such as DragGAN to enhance user control over the latent space. 

\subsection{DragGAN}
DragGAN is a technique designed for real-time, interactive manipulation of GAN-generated images by optimizing control points within the latent space. It allows users to define control points directly on images and apply localized edits, enabling precise adjustments of specific image attributes and intuitive modification. This achieved through a motion supervision loss function, which is critical in ensuring accurate spatial transformations in images. For further technical details, refer to \cite{Pan_2023}. For example, when applied to GAN-generated dog faces, DragGAN can adjust the pose of a subject while preserving fine-grained details like fur texture or facial expressions. This capability is particularly effective when combined with StyleGAN, offering a more accessible and intuitive way to control fine-grained image details.


However, while DragGAN improves control, the computational demands of working with high-dimensional latent spaces, such as the $W+$ space in StyleGAN, remain high. To streamline this process and improve efficiency, we integrate PCA-based dimensionality reduction, which allows a focus on the most significant variations within the latent space, potentially reducing computational costs and enhancing usability.

\subsection{Dimensionality Reduction}
Principal Component Analysis (PCA) is a dimensionality reduction technique that identifies the major directions of variance within data, simplifying complex spaces by focusing on their principal components. PCA measures the proportion of the dataset's variance captured by the principal components in PCA. 




This technique has been effectively applied in methods like GANSpace \cite{härkönen2020ganspacediscoveringinterpretablegan}, where PCA reduces the latent space of GANs to a lower-dimensional, more interpretable representation. By applying PCA to StyleGAN’s $W^+$ space, we aim to achieve efficient and disentangled image manipulations. Dimensionality reduction not only decreases the number of variables for optimization but also accelerates convergence, improving the speed and stability of latent space manipulations. This enables us to focus on primary components, creating a streamlined and interpretable space for precise and efficient GAN-based image editing.

\subsection{Latent Space Alignment}

Latent space alignment between GAN models involves mapping semantic attributes in the models' latent spaces to facilitate cross-domain semantic consistency in image manipulations. For example, if a change is made to an image of a dog generated by model $G_1$ that is pre-trained on AFHQ Dog (e.g. by changing its fur color or adding a smile), it might be a desired effect to replicate that change in an image of a cat generated by model $G_2$ that is pre-trained on AFHQ-Cat. There are several techniques used for alignment of GAN models, including: 

\begin{itemize}[noitemsep, , topsep=0.5pt]
\item GAN inversion and projection: GAN inversion involves projecting a generated image $I_1$ from the latent space of one GAN model ($G_1$) into the latent space of another GAN model ($G_2$). The goal is to find the closest latent vector $z_2$ such that $G_2(z_2) \approx I_1$. This aligns the two models by finding latent representations in one model that approximate outputs of the other \cite{xia2022ganinversionsurvey}.

\item Cross-model latent manipulation (with PCA): PCA is applied to the latent spaces of both $G_1$ and $G_2$ to identify the most significant latent directions related to key semantic transformations of interest (e.g., color or pose change). For cross-model manipulations, these latent directions are traversed in $G_1$ (e.g., modifying the fur color of a generated dog image from white to black). The  corresponding change is then made in $G_2$ to observe if it has a consistent effect (e.g, white cat becomes black cat) \cite{härkönen2020ganspacediscoveringinterpretablegan}. This qualitative evaluation offers insight into the semantic consistency between latent space manipulations of cross-domain models and can be integrated with techniques like DragGAN to achieve better control and interpretability in tasks like image synthesis, cross-domain translation and editing.
\end{itemize}

\section{Approach}

This work integrates several techniques to achieve precise and interpretable manipulations within StyleGAN's latent space. Our approach involves fine-tuning a pre-trained GAN on the AFHQ-Dog dataset, applying Principal Component Analysis (PCA) for dimensionality reduction, and leveraging the DragGAN framework for efficient latent space control. The methodology addresses cross-model latent space alignment and introduces robust evaluation metrics for analyzing the efficacy of latent manipulations. The core experimental steps are as follows:
\begin{enumerate}[noitemsep,, topsep=0.5pt]
    \item PCA Reduction Setup: Configure PCA reduction with StyleGAN for dimensionality control and model switching across various datasets.
    \item Hyperparameter Tuning: Experimentation with style layers, principal components, and learning rates to study their effects on performance and control.
    \item Latent Vector Manipulation: Targeted manipulation of latent vectors, evaluating precision and consistency of results across different datasets.
\end{enumerate}
The main metrics used in the analysis are:
\begin{itemize}[noitemsep, topsep=0.5pt]
    \item Optimization Time: By adjusting hyperparameters such as learning rate and number of principal components, we evaluated and recorded the effects on processing time per iteration.
    \item Motion Loss: Provides insight into convergence behavior and stability, especially crucial when incorporating PCA reduction.
    \item Number of Iterations (specific to DragGAN): Counted to quantify the accuracy of direct latent space manipulations.
    \item The Structural Similarity Index Measure (SSIM): Evaluates image similarity by considering luminance, contrast, and structural information. It is a robust image-specific measure commonly applied to compare and contrast images, see \cite{wang_ssim} for more details.


\end{itemize}

Our implementation was built upon publicly available StyleGAN \cite{karras2020analyzingimprovingimagequality} and DragGAN \cite{Pan_2023} repositories. Notable modifications include: integration of PCA reduction within the original DragGAN framework, custom modules for GAN inversion and cross-model latent mapping, and extensions to TensorBoard logging for detailed metrics tracking.



\section{GAN Fine-Tuning}

\begin{figure}[htb!]
    \centering
    \captionsetup{width=0.90\linewidth}
    \includegraphics[width=0.90\linewidth]{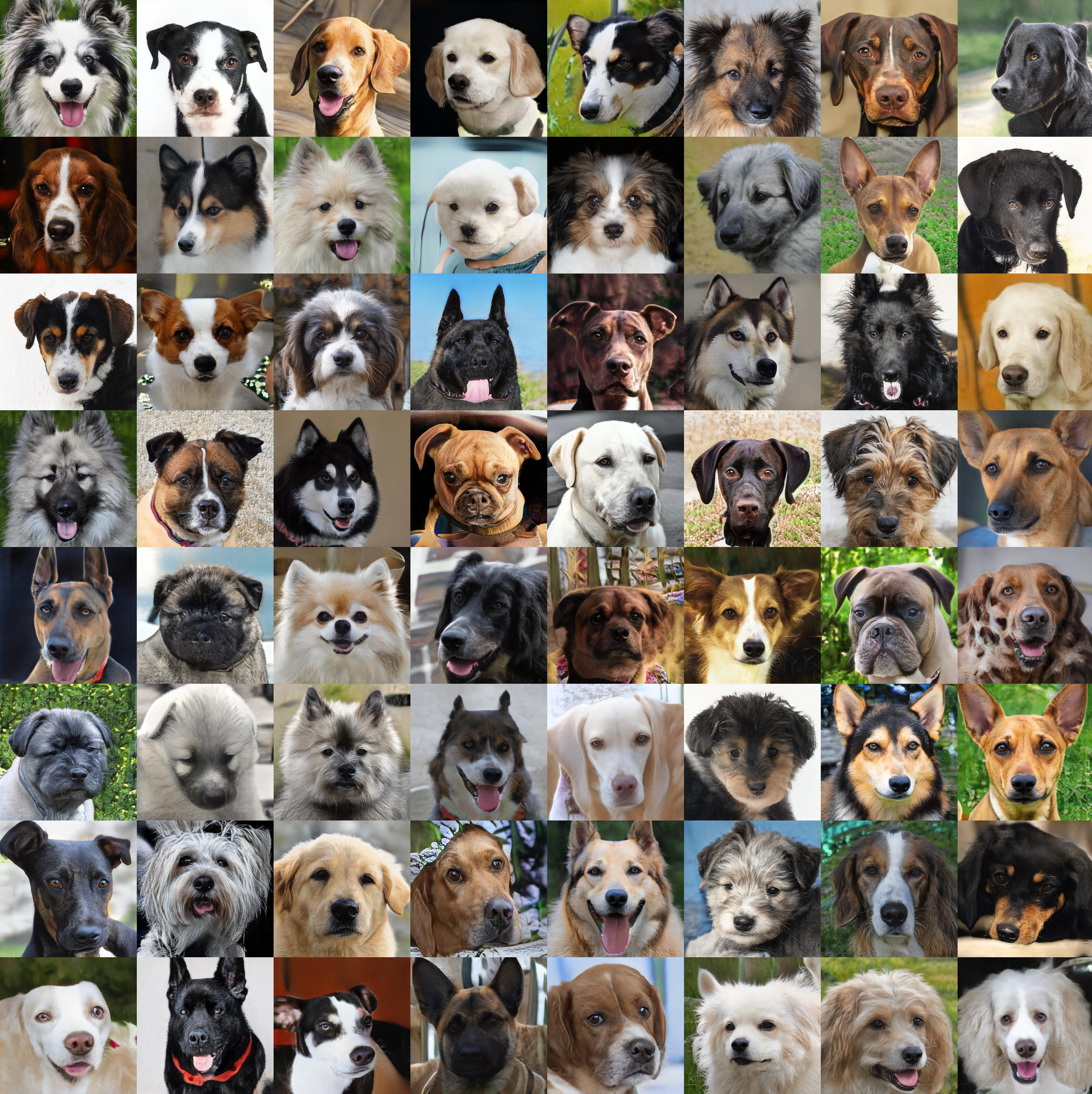}
    \caption{Randomly sampled examples from our fine-tuned StyleGAN2 generator model.}
    \label{fig:stylegan_examples}
    \vspace{-12pt}
\end{figure}

As an initial starting point, we utilized the FFHQ (human faces) StyleGAN pre-trained model as a baseline and fine-tuned it on the AFHQ Dog dataset. The fine-tuning process demonstrated the effectiveness of transfer learning in the StyleGAN framework (see Figure \ref{fig:stylegan_examples} for some images our model generated).

\renewcommand{\arraystretch}{1.0}
\begin{table}[htb]
\centering
\captionsetup{width=0.95\linewidth}
\footnotesize
\resizebox{\columnwidth}{!}{%
\begin{tabular}{cccccc}
\hline
Models & Pretrained & kimgs & nGPUs & FID   & Wallclock \\ \hline
Baseline & scratch    & 25k                 & 8     & 19.37 & -              \\
ADA        & scratch    & 25k                 & 8     & 7.4   & -              \\ \hline
ADA  & FFHQ       & 145                   & 1     & 17.39 & 10h 9m     \\ \hline
\end{tabular}%
}
\vspace{-5pt}
\caption{Summary of our fine-tuned StyleGAN2 model as compared to the results obtained in \cite{karras2020analyzingimprovingimagequality}.}
\label{tab:finetune_summary}
\vspace{-5pt}
\end{table}

\begin{figure}[htb!]
    \centering
    \captionsetup{width=0.95\linewidth}
    \includegraphics[width=0.85\linewidth]{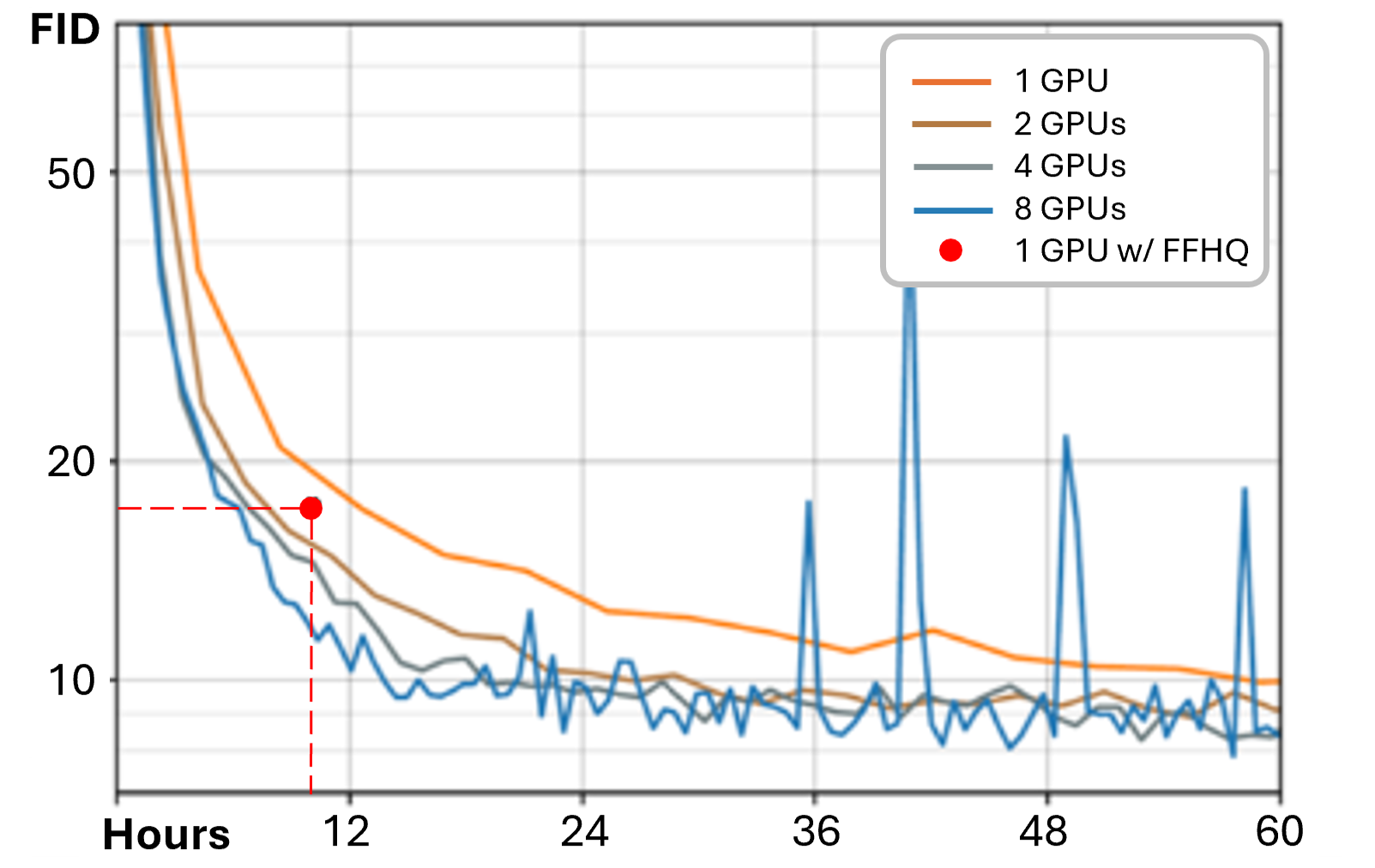}
    \caption{Recreated convergence curves for the AFHQ dataset from scratch as a function of wallclock time, compared with our fine-tuned implementation using pre-trained FFHQ model.}
    \label{fig:finetuned_loss}
    \vspace{-5pt}
\end{figure}

To evaluate the benefits of fine-tuning, we compared the Fréchet Inception Distance (FID) scores of the fine-tuned model with those of similar models trained from scratch on the same dataset. The comparison result are summarized in Table \ref{tab:finetune_summary}. When considering the same hardware (1 GPU), the fine-tuned model outperformed the Baseline from-scratch models, achieving improved FID scores. However, when looking at the equivalent ADA architecture, there is still much to be gained if we were to continue training on more images for longer, as evidenced in Figure \ref{fig:finetuned_loss}.

The computational time required for a competitive fine-tuned model remains a key limitation. While a single fine-tuning took approximately 10 hours on our system, the best models (ADA), trained from scratch required significantly longer, typically 3 days. To ensure a focus on the interpretability and reproducibility of latent space manipulations, we chose to use NVIDIA's pre-trained model for the subsequent experiments, rather than our fine-tuned model.

\section{Experiments and Results}
All experiments were systematically logged using TensorBoard~\cite{tensorflow2015-whitepaper} to ensure a consistent and rigorous evaluation. Key metrics, such as optimization time, loss function values, gradient magnitudes, and handle-point distances were tracked across iterations to assess the model’s performance and efficiency under various hyper-parameter configurations

Table \ref{tab:gridsearch_summary} presents the range of hyperparameters evaluated, including learning rates, the number of PCA components, and layer-specific settings. The table summarizes the grid search conducted, ensuring reproducibility and clarity in experimental design. These variables were chosen to optimize the balance between computational efficiency and image quality.

\renewcommand{\arraystretch}{1.0}
\begin{table}[htb]
\centering
\captionsetup{width=0.9\linewidth}
\footnotesize
\begin{tabular}{p{0.45\columnwidth}  p{0.45\columnwidth}}
\hline
Hyperparameter      & Value                   \\ \hline
Learning Rate       & \{0.1, 0.05, 0.002\}    \\
Component Number    & \{64, 128, 256, 512\}   \\
$W^+$ Layer Number    & \{3, 6, 12\}            \\
Stopping Distance   & \{10 pixels\}           \\
Max Iterations      & \{150\}                 \\
PCA Samples         & \{1000\}                \\
Optimizer           & \{AdamW\}               \\ 
Seeds               & \{13, 42, 999\}        \\   \hline
\end{tabular}
\vspace{-5pt}
\caption{Summary of grid search hyper-parameters}
\label{tab:gridsearch_summary}
\vspace{-5pt}
\end{table}

\subsection{Latent Space Dimensionality Reduction}
\begin{figure}[htb]
    \centering
    \captionsetup{width=0.95\linewidth}
    \includegraphics[width=\linewidth]{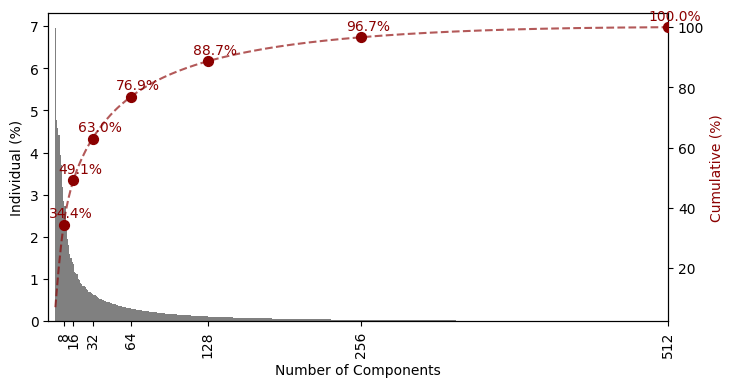}
    \caption{Explained variance of each principle component within the reduced dimensional $W^+$-space}
    \label{fig:pca_distribution}
    \vspace{-10pt}
\end{figure}

Prior to conducting experiments, a PCA model was applied to the $W^+$ latent space layers of the StyleGAN model. The cumulative explained variance across the principal components was plotted within Figure \ref{fig:pca_distribution}, revealing the trade-off between dimensionality reduction and total variance captured for each specific layer. 

Retaining fewer components reduces the number of optimization parameters, enhancing computational efficiency, but sacrifices the representation of some variance in the data. An interesting observation was that the distribution ratios remained consistent regardless of the number of layers included in the PCA reduction. While the magnitudes of variance captured varied, they were proportional to the number of layers included. This suggests that each StyleGAN layer progressively encodes features, building complexity as training progresses, consistent with the progressive training methodology of StyleGAN.

\subsection{DragGAN Experiments}

\begin{figure}[th!]
    \centering
    \captionsetup{width=0.95\linewidth}
    \includegraphics[width=0.7\linewidth]{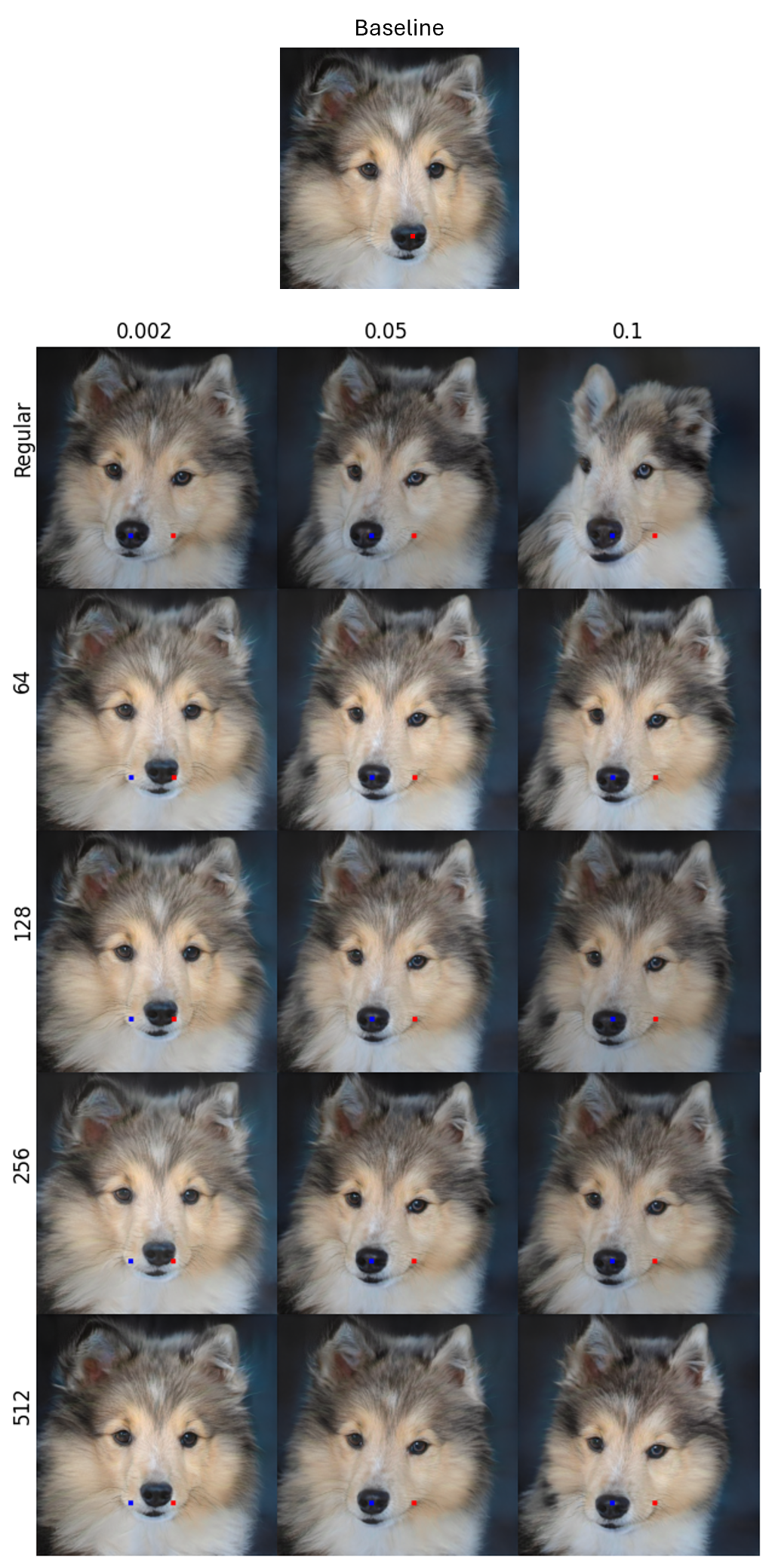}
    \caption{Results of PCA applied to $W^+\text{ layers}=3$ during the DragGAN process. Rows correspond to different number of principle components retained and columns represent the varying learning rates. The blue dots indicate handle points, while the red dots mark the target points. Note that the left column (learning rate of 0.002) using PCA did not converge to the goal.}
    \label{fig:generated_images}
    \vspace{-12pt} 
\end{figure}

The experimental results provide an overview of the optimization efficiency and stability achieved with PCA-enhanced setups across various hyperparameter configurations. Qualitative evaluations using visual comparisons of generated images are shown in Figure \ref{fig:generated_images}, which includes results for the first three StyleGAN $W^+$ layers with handle and target points (additional cases in Appendix \ref{app:addimages}). Quantitative summaries of global optimization results are presented in Table \ref{tab:comparison}, with metrics like average optimization time and loss stability visualized in Figure \ref{fig:individual_optimizers} (complete results in Appendix \ref{app:completeresults}). Notably, observations from initial tests align closely with extended test results, highlighting the following insights:

\begin{figure*}[ht!]
    \captionsetup{width=0.95\linewidth}
      \includegraphics[width=\textwidth]{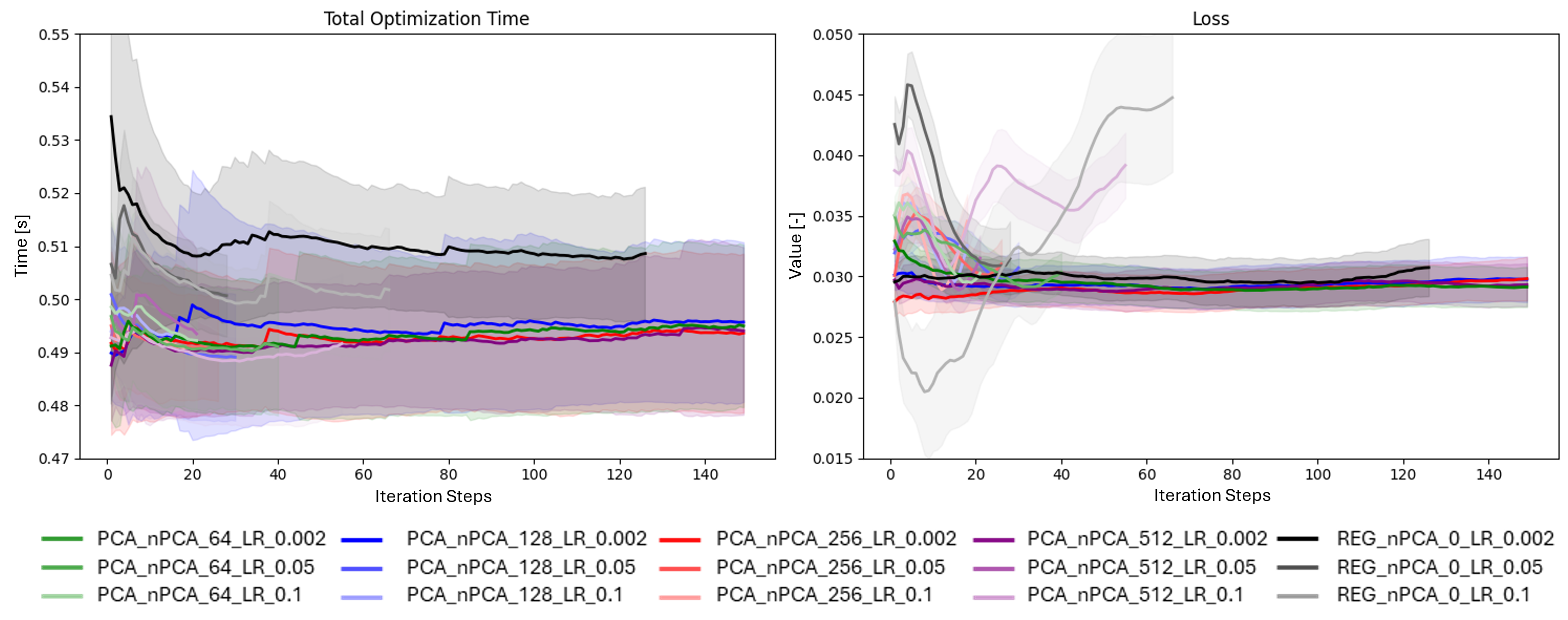}
    \vspace{-15pt}
    \caption{Average ($\pm1\sigma$) performance curves when considering $W^+$ layers$=3$ for total optimization time and motion loss metrics with applied smoothening ($\alpha=0.99$) to better observe trends.}
    \label{fig:individual_optimizers}
    \vspace{-10pt}
\end{figure*}

\begin{enumerate}[noitemsep, topsep=0.5pt]
    \item The total optimization time per iteration consists of the optimization-step and point-tracking phases. Surprisingly, regular DragGAN models are slightly faster during the optimization-step phase due to the minor overhead introduced by PCA transformations. However, PCA models introduce improvement ($\sim4\%$) in point-tracking time, despite this phase operating on non-dimensionalized images. This suggests that PCA indirectly stabilizes the optimization trajectory, improving initialization or context for point-tracking.
    \item At smaller learning rates, regular models show highly unstable results. While PCA-reduced variants also show some instability, they are better at avoiding severe impacts and retain a high degree of image quality. However, many PCA gradients approach instability if convergence is not quickly achieved. 
    
\end{enumerate}
The integration of PCA dimension reductions into the latent W+ layers of DragGAN demonstrates a clear trade-off between computational efficiency and output quality. Based on the global performance metrics in Table \ref{tab:comparison}, the following conclusions can be drawn:
\begin{enumerate}[noitemsep, topsep=0.5pt]
    \item PCA consistently reduces total optimization time ($t_{total}$) across all configurations, demonstrating its value in enhancing computational efficiency. The time savings, particularly in the SSIM-$time_{total}$ trade-off, highlights PCA’s ability to make the optimization process faster without significant performance losses.
    \item Shallower latent spaces (W+ layers=3 \& 6) achieve higher SSIM values than deeper layers. PCA reductions are particularly effective for these configurations, improving efficiency with minimal quality degradation. Even with smaller PCA components (e.g., 64), quality remains comparable to the baseline, highlighting PCA's potential to streamline optimization.
    \item PCA-reduced models struggle to converge within the iteration cap (150) at lower learning rates (0.002), see Figure \ref{fig:generated_images}. This may be due to inherent noise in the reduced latent space, which complicates optimization with small step sizes. Higher learning rates stabilize the process and enable faster convergence, suggesting that the PCA-reduced loss space may not be entirely smooth and more suited for larger step sizes. 
    
\end{enumerate}

\vspace{-5pt}
\renewcommand{\arraystretch}{0.95}
\begin{table}[t!]
\captionsetup{width=0.95\linewidth}
\centering
\footnotesize
\begin{tabular}{>{\centering\arraybackslash}p{1.5cm} >{\centering\arraybackslash}p{2.5cm}ccccccccc}
\toprule
\multirow{2}{*}{\textbf{nPCA}} &  \multirow{2}{*}{\textbf{Metric}} & \multicolumn{3}{c}{\textbf{$W^+$ layers = 3}} \\
\cmidrule(lr){3-5} 
& & \textbf{0.002} & \textbf{0.05} & \textbf{0.1}  \\ 
\midrule
\multirow{2}{*}{Regular} & $\text{iteration}_{\text{total}}$ & 126 & 17 & 44 \\
                         & SSIM/time \scriptsize{($10^{\text{-}2}$)}  & \cellcolor[HTML]{F8716C}1.054 & \cellcolor[HTML]{9ACE7F}5.71 & \cellcolor[HTML]{FDCC7E}2.323 \\
\midrule
\multirow{2}{*}{64} & $\text{iteration}_{\text{total}}$ & 150 (\scriptsize{\xmark}) & 40 & 22  \\
                    & SSIM/time \scriptsize{($10^{\text{-}2}$)} & \cellcolor[HTML]{F8716C}1.054 & \cellcolor[HTML]{EEE683}3.271 & \cellcolor[HTML]{99CE7F}5.749 \\

\midrule
\multirow{2}{*}{128} & $\text{iteration}_{\text{total}}$ & 150 (\scriptsize{\xmark}) & 31 & 19 \\
                     & SSIM/time \scriptsize{($10^{\text{-}2}$)} & \cellcolor[HTML]{F8716C}1.037 & \cellcolor[HTML]{CDDD82}4.238 & \cellcolor[HTML]{75C47D}6.795 \\
\midrule
\multirow{2}{*}{256} & $\text{iteration}_{\text{total}}$ & 150 (\scriptsize{\xmark}) & 27 & 19 \\
                     & SSIM/time \scriptsize{($10^{\text{-}2}$)} & \cellcolor[HTML]{F8716C}1.032 & \cellcolor[HTML]{B5D680}4.936 & \textbf{\cellcolor[HTML]{6EC17C}7.009}  \\
\midrule
\multirow{2}{*}{512} & $\text{iteration}_{\text{total}}$ & 150 (\scriptsize{\xmark}) & 22 & 49  \\
                     & SSIM/time \scriptsize{($10^{\text{-}2}$)} & \cellcolor[HTML]{F8716C}1.010 & \cellcolor[HTML]{91CC7E}5.986 & \cellcolor[HTML]{FEE182}2.619\\

\bottomrule
\end{tabular}
\caption{Averaged results across 3 different seeds for different number of principle components (nPCA) and learning rates. The SSIM/time ratio is used as a measure of success, where this ratio captures the trade-off between achieving high image similarity and minimizing evaluation time. Higher values (green) indicate better performance, balancing quality and efficiency. Non-converged solutions are indicated as ({\scriptsize{\xmark}}).}
\label{tab:comparison}
\vspace{-10pt}
\end{table}




\subsection{Latent Space Alignment}
\begin{figure}[htb!]
    \centering
    \captionsetup{width=0.95\linewidth}
    \includegraphics[width=\linewidth]{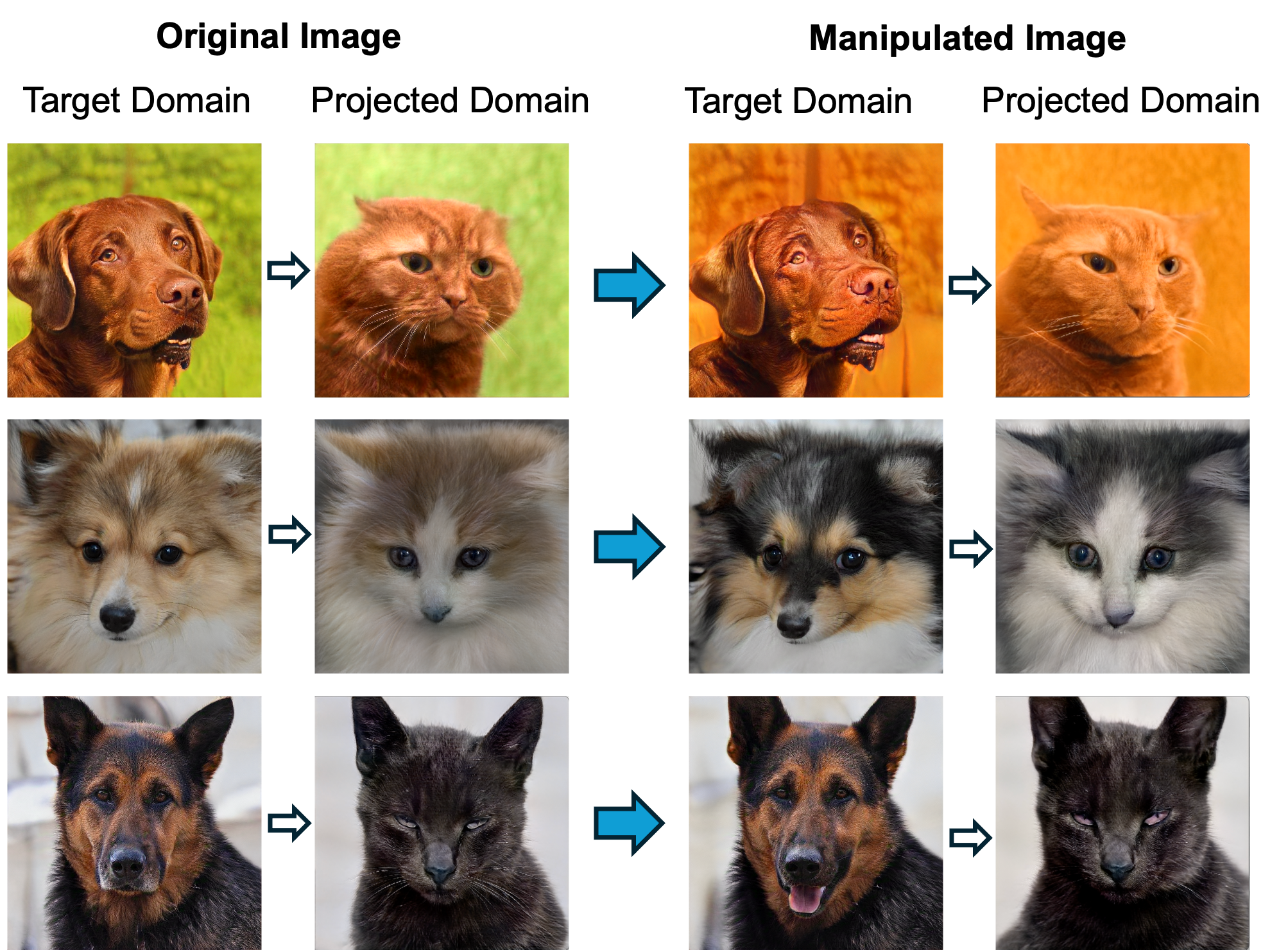}
    \caption{Alignment between two GAN models $G1$ and $G2$, trained on AFHQ-Dog and AFHQ-Cat respectively. Images are first generated by $G1$ and then projected into the W-space of $G2$ \textit{(left)}. The original $G1$ images are manipulated along key latent vectors to induce edits; these manipulated images are subsequently projected into $G2$ \textit{(right)}. While high-level edits to background and color are captured, finer details of facial features and expressions are not retained by this method.}
    \label{fig:gan_inversion}
    \vspace{-12pt}
    \end{figure}

DragGAN offers one possible technique for image synthesis and editing by manipulating the generative latent space. Controlling the latent space and aligning generative outputs between different domains could be desirable extensions to DragGAN or similar frameworks, and requires specialized optimization techniques to ensure consistency and quality. Here, we conduct a preliminary exploration of image editing and latent space alignment techniques, drawing on methods from StyleGAN and GANSpace. Figure \ref{fig:gan_inversion} shows sample images generated by these methods. Images in the initial (target) domain, AFHQ-Dog, are projected into the latent space of AFHQ-Cat, to create corresponding cat images that retains key features of the target dog image, such as background, pose and fur color, while being visually recognizable as cats.

Despite the limitation of aligning separate pre-trained models that do not share a common mapping in $W/W^+$ latent space, the projected images are of high visual quality and retain many high-level features from the target image. Future work could involve training a dedicated encoder for cross-domain image embedding, facilitating more efficient optimization of the projected images.

We also explore image editing using interpretable latent control as implemented in GANSpace \cite{härkönen2020ganspacediscoveringinterpretablegan}. Starting with a target domain image (dog), we apply edits by moving it along key latent dimensions identified by PCA. These edits include changes to the background color, fur color and the addition of a smile. Because the domains are not mapped to a common latent space, we attempt alignment via projection again, by projecting the manipulated dog image into AFHQ-Cat latent space to produce a corresponding, manipulated cat image. While this projection captures higher-level manipulations like background and fur color, it struggles with finer modifications to facial expression and details. Training a unified model to map both domains to a shared latent representation could address this limitation. For example, CycleGANs have demonstrated strong performance in unpaired, cross-domain image translation tasks \cite{zhu2017unpaired}.

\section{Conclusions and Future Direction}

The integration of PCA-based dimensionality reduction of latent space into the DragGAN image manipulation framework provides notable efficiency gains while maintaining a high degree of structural information and preserving image quality. However, utilizing PCA in the DragGAN framework also introduces challenges related to optimization stability and convergence. Therefore, we suggest the following future work:

\begin{enumerate}[noitemsep, topsep=0.5pt]
    \item Developing hybrid learning rate schedules to balance initial convergence speed with improved stability during local refinements.
    \item Conducting an in-depth analysis of the PCA-reduced loss landscape to better understand how dimensionality reduction influences optimization dynamics.
    \item Investigating potential noise in PCA-reduced latent spaces and its impact on the optimization trajectory.
\end{enumerate}

Overall, our findings suggest that PCA-enhanced workflows not only streamline optimization but also allow for stable and interpretable alignment across diverse latent spaces, as demonstrated by visually aligning the outputs of GANs trained on the AFHQ-Dog and AFHQ-Cat datasets. Tying this capability into PCA-reduced DragGAN could facilitate image editing tools well suited for cross-domain image synthesis and editing tasks, as well as applications requiring efficient or interpretable latent space navigation in constrained computational settings.

\raggedbottom


\appendix



\section{Complete Images}\label{app:addimages}

\begin{figure*}[t!]
    \centering
    \begin{subfigure}[t]{0.48\textwidth} 
        \centering
        \includegraphics[width=\textwidth]{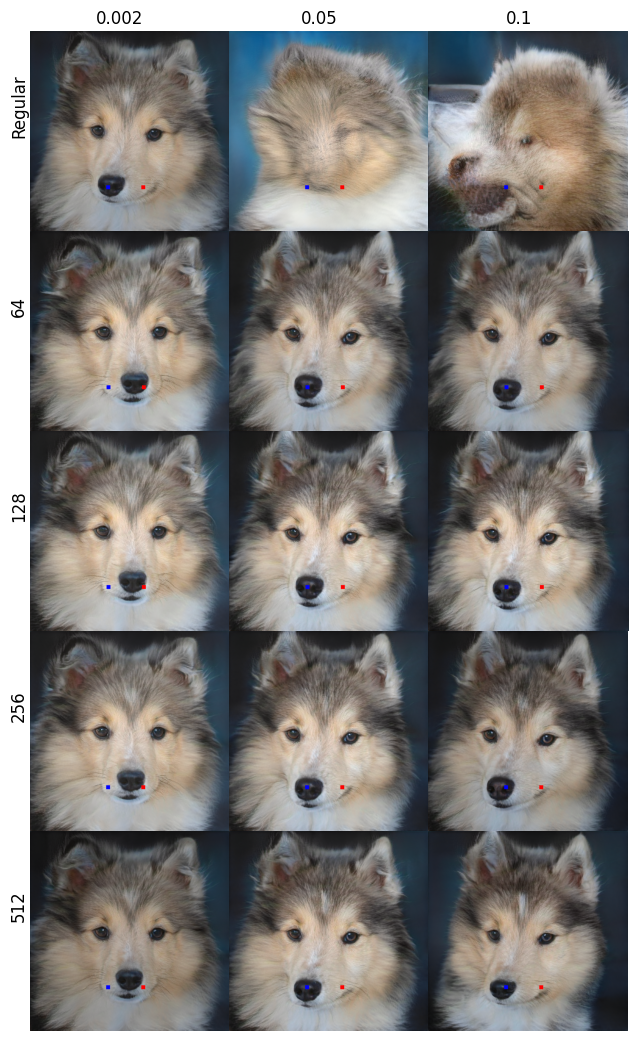}
        \caption{$W^+\text{ layers}=6$.}
        \label{fig:appendix_w6}
    \end{subfigure}
    \hfill
    \begin{subfigure}[t]{0.48\textwidth}
        \centering
        \includegraphics[width=\textwidth]{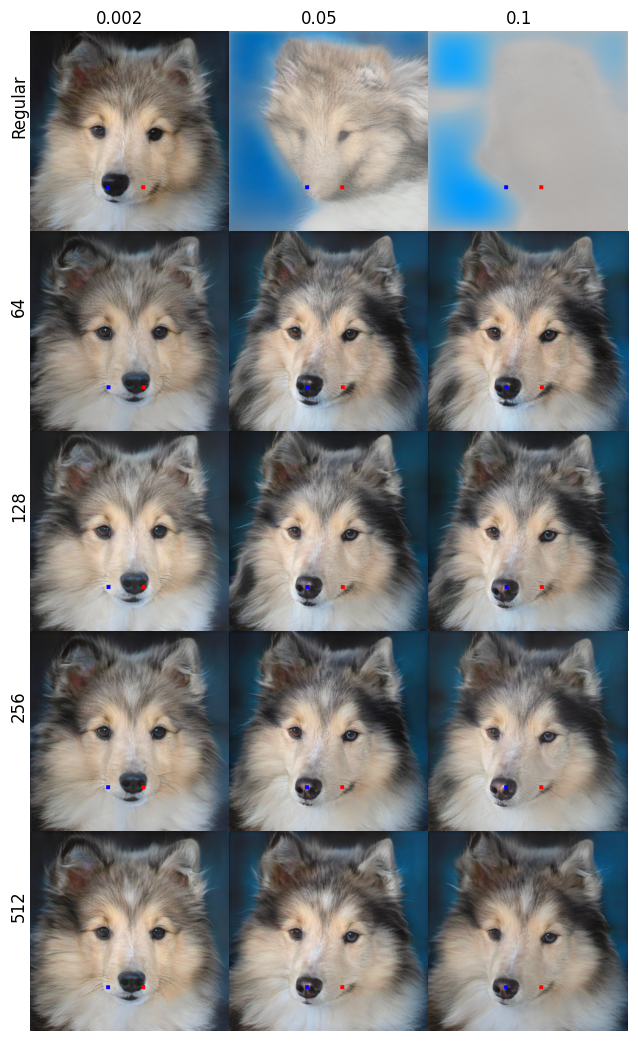}
        \caption{$W^+\text{ layers}=12$.}
        \label{fig:appendix_w12}
    \end{subfigure}
    \caption{Image results of PCA applied during the DragGAN process.}
    \label{fig:appendix-images}
\end{figure*}

Additional visualizations to further illustrate the results of applying PCA during the DragGAN process. Figures \ref{fig:appendix_w6} and \ref{fig:appendix_w12} showcase the completed image reconstructions for two configurations of $W^+$ layers, highlighting the effects of varying layer depths. These supplementary images offer a comprehensive view of the consistency and quality of the generated outputs across different settings.

\section{Complete Results}\label{app:completeresults}

\begin{figure*}[t!]
    \centering
    \captionsetup{width=0.95\linewidth}
      \includegraphics[width=\textwidth]{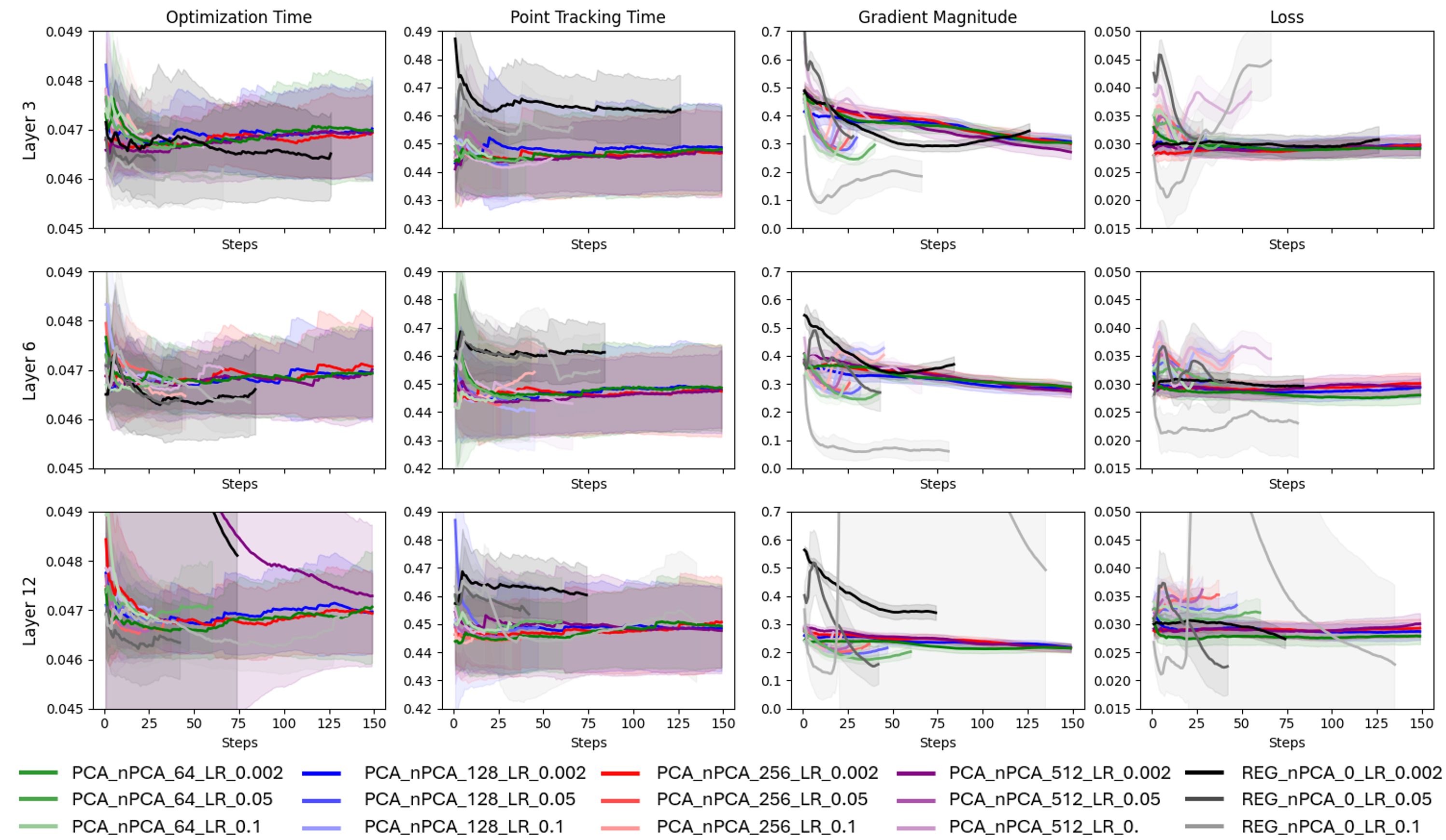}
    \caption{Average ($\pm1\sigma$) performance curves across all tested layers for optimization time, point-tracking time, gradient magnitude, and motion loss metrics with applied smoothening ($\alpha=0.99$). }
    \label{fig:appendix_individual_optimizers}
    \vspace{-2pt}
\end{figure*}

Additional quantitative results to further illustrate the results of applying PCA during the DragGAN process. Figure \ref{fig:appendix_individual_optimizers} and Table \ref{tab:appendix_summary_comparison} present the complete results for all configurations of $W^+$ layers, highlighting and comparing the effects of varying layer depths for all considered metrics.

\section{Available Repository}\label{app:repo}

All code and supplementary materials used in this research are publicly available in our GitHub repository. This includes scripts for pre-processing, model training, and latent space manipulation, as well as configurations for reproducing the results. The repository can be accessed at:
\url{https://github.gatech.edu/nkaushik31/DragGAN-Space.git}

\raggedbottom 
\onecolumn

\clearpage 

\renewcommand{\arraystretch}{1.1}
\FloatBarrier
\suppressfloats[b]
\begin{table*}[b]  
\captionsetup{width=0.95\linewidth}
\centering
\footnotesize
\begin{tabular}{>{\centering\arraybackslash}p{1.5cm} >{\centering\arraybackslash}p{2.5cm}ccccccccc}
\toprule
\multirow{2}{*}{\textbf{nPCA}} &  \multirow{2}{*}{\textbf{Metric}} & \multicolumn{3}{c}{\textbf{$W^+$ layers = 3}} & \multicolumn{3}{c}{\textbf{$W^+$ layers = 6}} & \multicolumn{3}{c}{\textbf{$W^+$ layers = 12}} \\
\cmidrule(lr){3-5} \cmidrule(lr){6-8} \cmidrule(lr){9-11}
& & \textbf{0.002} & \textbf{0.05} & \textbf{0.1} & \textbf{0.002} & \textbf{0.05} & \textbf{0.1} & \textbf{0.002} & \textbf{0.05} & \textbf{0.1} \\
\midrule
\multirow{3}{*}{Regular} & $\text{iteration}_{\text{total}}$ & 126 & 17 & 44 & 85 & 35 & 55 & 74 & 33 & 95 \\
                         & $\text{time}_{\text{total}}$ & 44.652 & 7.780 & 18.959 & 30.226 & 13.501 & 22.339 & 26.552 & 12.686 & 41.882 \\
                         & SSIM & 0.470 & 0.444 & 0.440 & 0.496 & 0.384 & 0.413 & 0.491 & 0.401 & 0.387 \\  
                         & SSIM/time \scriptsize{($10^{\text{-}2}$)}  & \cellcolor[HTML]{F8726C}1.054 & \cellcolor[HTML]{9ACE7F}5.710 & \cellcolor[HTML]{FDCC7E}2.323 & \cellcolor[HTML]{FA9B74}1.640 & \cellcolor[HTML]{FCEB84}2.845 & \cellcolor[HTML]{FBAA77}1.848 & \cellcolor[HTML]{FBAA77}1.850 & \cellcolor[HTML]{F2E784}3.157 & \cellcolor[HTML]{F8696B}0.924 \\
\midrule
\multirow{3}{*}{64} & $\text{iteration}_{\text{total}}$ & 150 (\scriptsize{\xmark}) & 40 & 22 & 150 (\scriptsize{\xmark}) & 43 & 23 & 150 (\scriptsize{\xmark}) & 60 & 32 \\
                    & $\text{time}_{\text{total}}$ & 51.939 & 14.068 & 7.929 & 52.058 & 15.146 & 8.348 & 51.894 & 21.061 & 11.494 \\
                    & SSIM & 0.547 & 0.460 & 0.456 & 0.552 & 0.464 & 0.460 & 0.551 & 0.428 & 0.424 \\
                    & SSIM/time \scriptsize{($10^{\text{-}2}$)} & \cellcolor[HTML]{F8726C}1.054 & \cellcolor[HTML]{EEE683}3.271 & \cellcolor[HTML]{99CE7F}5.749 & \cellcolor[HTML]{F8726C}1.061 & \cellcolor[HTML]{F5E884}3.065 & \cellcolor[HTML]{A1D07F}5.513 & \cellcolor[HTML]{F8726C}1.063 & \cellcolor[HTML]{FCB77A}2.033 & \cellcolor[HTML]{DFE283}3.687 \\

\midrule
\multirow{3}{*}{128} & $\text{iteration}_{\text{total}}$ & 150 (\scriptsize{\xmark}) & 31 & 19 & 150 (\scriptsize{\xmark}) & 33 & 27 & 150 (\scriptsize{\xmark}) & 47 & 26 \\
                     & $\text{time}_{\text{total}}$ & 52.077 & 10.973 & 6.776 & 52.002 & 11.569 & 11.868 & 52.021 & 16.652 & 9.283 \\
                     & SSIM & 0.540 & 0.465 & 0.460 & 0.542 & 0.459 & 0.448 & 0.543 & 0.442 & 0.442 \\
                     & SSIM/time \scriptsize{($10^{\text{-}2}$)} & \cellcolor[HTML]{F8716C}1.037 & \cellcolor[HTML]{CDDD82}4.238 & \cellcolor[HTML]{75C47D}6.795 & \cellcolor[HTML]{F8716C}1.043 & \cellcolor[HTML]{D6DF82}3.969 & \cellcolor[HTML]{DCE182}3.776 & \cellcolor[HTML]{F8716C}1.044 & \cellcolor[HTML]{FEE482}2.655 & \cellcolor[HTML]{BBD881}4.760 \\
\midrule
\multirow{3}{*}{256} & $\text{iteration}_{\text{total}}$ & 150 (\scriptsize{\xmark}) & 27 & 19 & 150 (\scriptsize{\xmark}) & 27 & 28 & 150 (\scriptsize{\xmark}) & 38 & 23 \\
                     & $\text{time}_{\text{total}}$ & 51.895 & 9.492 & 6.759 & 51.984 & 9.614 & 12.063 & 52.018 & 13.267 & 8.133 \\
                     & SSIM & 0.536 & 0.469 & 0.474 & 0.548 & 0.466 & 0.456 & 0.551 & 0.464 & 0.467 \\
                     & SSIM/time \scriptsize{($10^{\text{-}2}$)} & \cellcolor[HTML]{F8706C}1.032 & \cellcolor[HTML]{B5D680}4.936 & \cellcolor[HTML]{6EC17C}7.009 & \cellcolor[HTML]{F8726C}1.054 & \cellcolor[HTML]{B8D780}4.852 & \cellcolor[HTML]{DCE182}3.783 & \cellcolor[HTML]{F8726C}1.059 & \cellcolor[HTML]{E6E483}3.501 & \cellcolor[HTML]{99CE7F}5.747 \\
\midrule
\multirow{3}{*}{512} & $\text{iteration}_{\text{total}}$ & 150 (\scriptsize{\xmark}) & 22 & 49 & 150 (\scriptsize{\xmark}) & 23 & 55 & 150 (\scriptsize{\xmark}) & 29 & 18 \\
                     & $\text{time}_{\text{total}}$ & 51.838 & 7.829 & 17.425 & 51.883 & 8.136 & 19.695 & 52.513 & 10.210 & 6.460 \\
                     & SSIM & 0.524 & 0.469 & 0.456 & 0.548 & 0.480 & 0.438 & 0.540 & 0.473 & 0.472 \\           
                     & SSIM/time \scriptsize{($10^{\text{-}2}$)} & \cellcolor[HTML]{F86F6C}1.010 & \cellcolor[HTML]{91CC7E}5.986 & \cellcolor[HTML]{FEE182}2.619 & \cellcolor[HTML]{F8726C}1.056 & \cellcolor[HTML]{93CC7E}5.905 & \cellcolor[HTML]{FCC57C}2.226 & \cellcolor[HTML]{F8706C}1.027 & \cellcolor[HTML]{BFD981}4.636 & \cellcolor[HTML]{63BE7B}7.305 \\
\bottomrule
\end{tabular}
\caption{Average optimization metrics across 3 different seeds for different nPCA, Learning Rates, and Number of Layers. Non-converged solutions are indicated as ({\scriptsize{\xmark}}).}
\label{tab:appendix_summary_comparison}
\end{table*}

\raggedbottom

\end{document}